\documentclass[fleqn,10pt]{wlscirep}
\usepackage[utf8]{inputenc}
\usepackage[T1]{fontenc}
\usepackage{siunitx}

\title{The Dresden Dataset for 4D Reconstruction of Non-Rigid Abdominal Surgical Scenes}

\author[1,*]{Reuben Docea}
\author[2,3,4]{Rayan Younis}
\author[5]{Yonghao Long}
\author[1]{Maxime Fleury}
\author[1]{Jinjing Xu}
\author[1,2]{Chenyang Li}
\author[3]{André Schulze}
\author[3]{Ann Wierick}
\author[1]{Johannes Bender}
\author[1]{Micha Pfeiffer}
\author[5]{Qi Dou}
\author[3,4$\dagger$]{Martin Wagner}
\author[1,4,$\dagger$]{Stefanie Speidel}

\affil[1]{Department of Translational Surgical Oncology, National Center for Tumor Diseases (NCT), NCT/UCC Dresden, a partnership between DKFZ, Faculty of Medicine and University Hospital Carl Gustav Carus, TUD Dresden University of Technology, and Helmholtz-Zentrum Dresden-Rossendorf (HZDR), Dresden, Germany}
\affil[2]{Department of Translational Surgical Oncology, NCT/UCC Dresden, Faculty of Medicine and University Hospital Carl Gustav Carus, TUD Dresden University of Technology, Fetscherstraße 74, 01307 Dresden, Germany}
\affil[3]{Department for Visceral, Thoracic and Vascular Surgery, Faculty of Medicine and University Hospital Carl Gustav Carus, Technische Universität Dresden, Dresden, Germany} 
\affil[4]{Centre for Tactile Internet with Human-in-the-Loop (CeTI), Technical University Dresden, Dresden, Germany}
\affil[5]{Department of Computer Science and Engineering, The Chinese University of Hong Kong, Hong Kong, China}

\affil[$\dagger$]{These authors contributed equally and jointly supervised this work.}
\affil[*]{corresponding author: Reuben Docea (reuben.docea@nct-dresden.de)}

\begin{abstract}
The \textbf{D4D} Dataset provides paired endoscopic video and high-quality structured-light geometry for evaluating 3D reconstruction of deforming abdominal soft tissue in realistic surgical conditions. Data were acquired from six porcine cadaver sessions using a da Vinci Xi stereo endoscope and a Zivid structured-light camera, registered via optical tracking and manually curated iterative alignment methods. Three sequence types - whole deformations, incremental deformations, and moved-camera clips - probe algorithm robustness to non-rigid motion, deformation magnitude, and out-of-view updates. Each clip provides rectified stereo images, per-frame instrument masks, stereo depth, start/end structured-light point clouds, curated camera poses and camera intrinsics. In postprocessing, ICP and semi-automatic registration techniques are used to register data, and instrument masks are created. The dataset enables quantitative geometric evaluation in both visible and occluded regions, alongside photometric view-synthesis baselines. Comprising over 300,000 frames and 369 point clouds across 98 curated recordings, this resource can serve as a comprehensive benchmark for developing and evaluating non-rigid SLAM, 4D reconstruction, and depth estimation methods.
\end{abstract}
\begin{document}

\flushbottom
\maketitle

\thispagestyle{empty}

\section*{Background \& Summary}
The ability to accurately reconstruct deforming soft tissue scenes is foundational to many computer-aided innovations in minimally invasive surgery (MIS). These methods are essential in particular for Image Guidance Systems (IGS), which are used for intraoperative navigation. Here, they provide a complete view of the surgical scene to improve spatial understanding, thereby supporting enhanced decision-making and safety\cite{wang2022endonerf, endosurf, schmidt2024tracking}. This improved guidance enables more precise instrument control and represents a key step towards robotic surgery automation\cite{wang2022endonerf, lu2021super, superdeep_2021}. For patients, these capabilities could lead to more effective interventions and may reduce the need for invasive follow-up procedures\cite{schneider2021igsreview}. Beyond the operating room, such reconstruction techniques are also highly valuable for creating realistic simulators for surgical education, training, and rehearsal.

However, IGS built on these principles have only seen scarce up-take for the case of abdominal surgery\cite{ciria2016comparative}. A primary reason for this is that the abdominal environment is non-rigid and continually subject to change. Designing a navigation system for such conditions is a considerable endeavour which requires a thorough understanding of system accuracy.

Soft tissue navigation typically relies upon the registration of preoperative data (such as CT or MRI scans) with intraoperative data, often consisting of a 3D map of the tissue surface\cite{thompson2018vivo_smartliver, heiselman2017characterization_explorerliver, docea2022laparoscopic}. These 3D maps, however, require time to create, and the methods to do so are often designed with the assumption that the scene is rigid. This means that as soon as the shape of the tissue changes, through manipulation from the surgeon or otherwise, it must be re-mapped for any subsequent registration to be accurate, which takes a substantial amount of time.

For this scenario, where tissue deforms even out of view\cite{azagra2023endomapper, docea2024seesaw}, a non-rigid mapping method is required. Such a method must realistically update previously mapped areas, which are no longer visible, to fit current observations. While there are works which go in this direction, there is no dataset available at present that serves to evaluate this capability. A first method which makes inroads into this problem is that of MIS-SLAM\cite{song2018mis}, which utilises an embedded deformation graph\cite{sumner2007embeddeddeformation} and a heuristics-based optimisation objective to update a global map from current observations. In a recent work\cite{docea2024seesaw}, we similarly attempted, using a GNN, to predict the positions of points which were seen but moved out of view, based on currently visible points.

In recent years, there have been many efforts at performing 3D reconstruction of non-rigid structures\cite{lin2022occlusionfusion, li20214dcomplete}. Of particular interest, these efforts span into the surgical domain, where soft tissue is concerned\cite{schmidt2024tracking}. Among such efforts are firstly some that pertain to single view-point 4D reconstructions, learning the deformations of specific scenes\cite{wang2022endonerf, endosurf, endo4dgs, liu2024endogaussian, chen2025surgicalgaussian, yang2024forplane, yang2024deform3dgs, xu2025t2gs}. Besides these, efforts to this end exist in the context of non-rigid SLAM\cite{song2018mis, lamarca2020defslam}. However, existing datasets (such as SCARED\cite{allan2021SCARED}, SERV-CT\cite{edwards2022serv}, EndoMapper\cite{azagra2023endomapper} and StereoMIS\cite{hayoz2023learning}) do not possess endoscopic video paired with high quality ground truth geometric data which can be used to evaluate 3D reconstruction in deforming soft tissue contexts. The authors of EndoNeRF remark, "Due to clinical regulation in practice, it is impossible to collect ground truth depth for numerical evaluation on 3D structures" \cite{wang2022endonerf}. This leads to the formerly mentioned 4D reconstruction methods evaluating with reference almost exclusively to photometric errors, such as: PSNR (Peak Signal to Noise Ratio)\cite{hore2010psnr}, SSIM (Structural Similarity Index Measure)\cite{wang2004ssim} and LPIPS (Learned Perceptual Image Patch Similarity)\cite{zhang2018lpips}.

The \textbf{D4D Dataset} fills a crucial gap by providing the first dataset containing endoscopic video paired with reference geometric data, acquired using a structured-light camera (for initial and final tissue states) in controlled manipulations of porcine cadaver abdominal soft tissue. The dataset comprises three types of sequences:

{\renewcommand{\theenumi}{\roman{enumi}}
 \renewcommand{\labelenumi}{(\theenumi)}
 \begin{enumerate}
 \item \textbf{Whole deformation}: complete manipulation sequences (pushing or pulling).
 \item \textbf{Incremental deformation}: a manipulation divided into smaller increments, allowing detailed analysis of the tissue’s deformation process.
 \item \textbf{Moved camera}: an action performed in two halves – first with a static camera, then after the camera has been repositioned. This setup enables evaluation of how accurately structures that deform out of view can be reconstructed realistically.
 \end{enumerate}}

Because the endoscope and structured-light camera occupy distinct viewpoints, the dataset facilitates quantitative evaluation of geometric reconstruction accuracy in both visible and occluded regions in 3D space. Consequently, it provides a valuable resource for developing and assessing non-rigid SLAM and reconstruction algorithms under realistic, deformable surgical conditions. An accompanying project page can be found at: \url{https://reubendocea.github.io/d4d/}.

\section*{Methods}

\subsection*{Data Collection}

\subsubsection*{Experimental Setup}
\begin{figure}[ht]
\centering
\includegraphics[width=\linewidth]{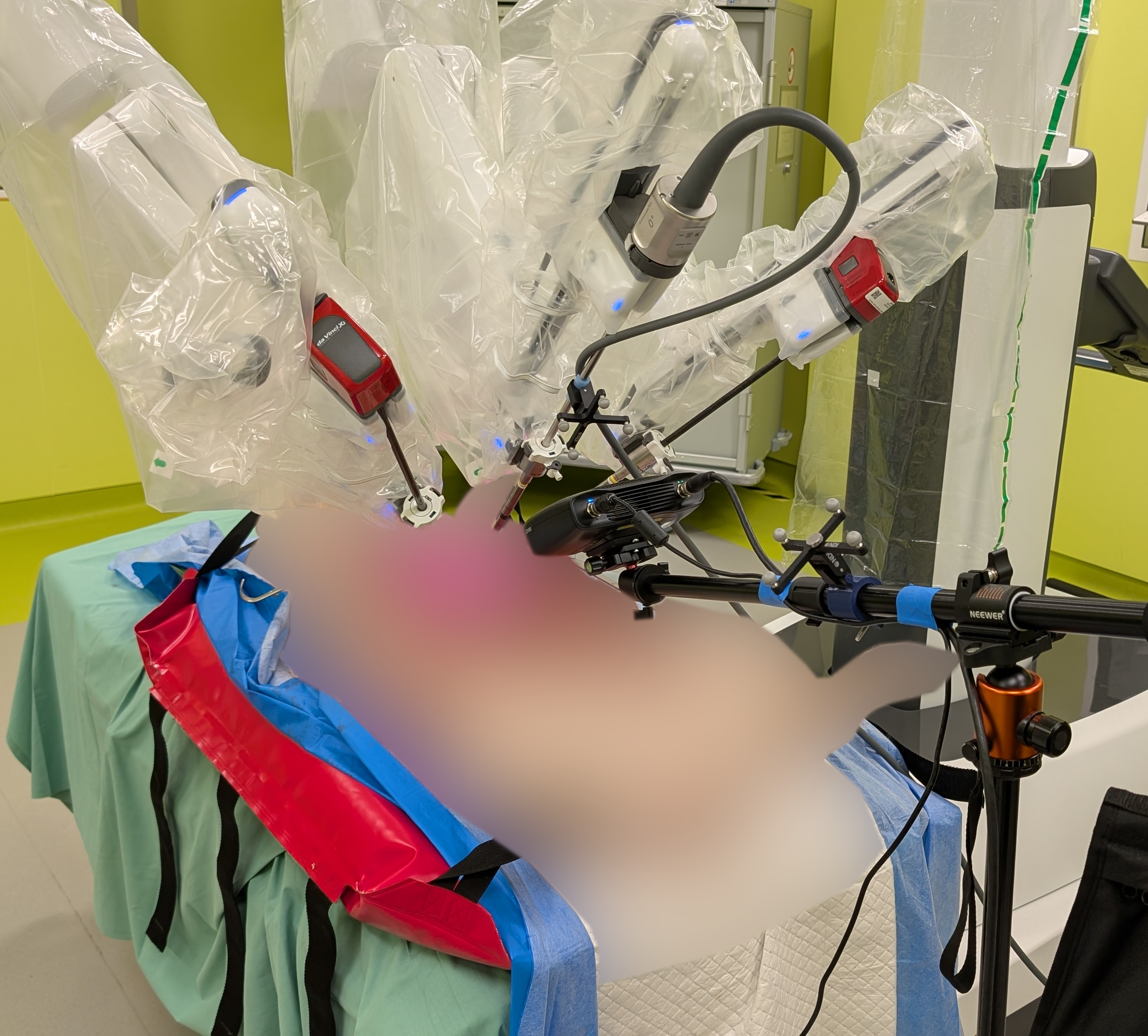}
\caption{Experimental setup showing porcine cadaver in supine position for open surgery with \textit{da Vinci}. The \textit{Zivid} structured light camera is visible in the foreground, with optical markers attached to the boom supporting the \textit{Zivid} and the endoscope.}
\label{fig:ExperimentalSetup}
\end{figure}

Between February 2025 and September 2025, data were collected using six porcine cadavers at the Experimental Operating Room of the \textit{National Center for Tumor Diseases} in \textit{Dresden, Germany}. All procedures were carried out with a \textit{da Vinci Xi} surgical robot (Intuitive Inc., Sunnyvale, USA) equipped with a stereo endoscope, using either a 30${^\circ}$ or a 0${^\circ}$ camera tip. Both left and right endoscopic images were recorded at a resolution of $894 \times 714$ pixels. To capture complementary geometric information, a Zivid 2+ M60 (Zivid AS, Oslo, Norway) structured light camera (SLC) was employed. Both the endoscope and the SLC were rigidly fitted with optical fiducial markers, whose poses were tracked in real time by a Polaris optical tracking system (Northern Digital Inc., Ontario, Canada). This setup enabled accurate registration between the visual and geometric data streams (Figure~\ref{fig:ExperimentalSetup}). The endoscope, SLC and Polaris data streams were recorded simultaneously on a workstation with Robot Operating System 2 (ROS2\cite{macenski2022ros2}), as \textit{rosbags}. Synchronisation was achieved via timestamping of all messages using the ROS2 system clock, ensuring temporal alignment across data streams. On some of the days of the data collection sessions, the relevant anatomical structures of the pig which are relevant to this study were operated on in the prior procedures. 

\subsubsection*{Experimental Protocol}
The stereo endoscope first underwent standard camera calibration following Zhang’s method~\cite{zhang2000flexible}, and hand–eye calibrations were performed to relate the poses of both the endoscope and the SLC to their respective tracking markers. Following calibration and conversion to open surgery, the data collection began. Room lighting was turned off at the beginning of data collection to enhance SNR of SLC and to better emulate lighting conditions of MIS.

A surgeon operating the \textit{da Vinci} system was asked to manipulate tissues in the abdomen from multiple viewing angles. These manipulations consisted primarily of pushing and pulling motions, repeated several times over the course of each session. Each segment, which we also refer to as \textit{clips}, followed a consistent structure: with the scene static and the endoscope light off, the SLC acquired a point cloud representing the baseline geometry; next, the endoscope light was switched on and the surgeon performed a tissue manipulation; finally, with the scene static again, the endoscope light was turned off again and a second SLC point cloud was captured to represent the post-action state. This alternating cycle ensured paired visual–geometric records of tissue deformation. As previously mentioned, three types of recording were performed: \textbf{(i)} whole deformations, representing complete push/pull actions; \textbf{(ii)} incremental deformations, divided into smaller steps to capture intermediate tissue states; and \textbf{(iii)} moved-camera sequences, where manipulation was split before and after camera repositioning to assess reconstruction continuity for structures moving out of view.

\subsection*{Ethics Statement}
All animals were euthanised at the conclusion of the prior procedures (which took place immediately before data collection sessions for this study) while still under deep anesthesia with the intravenous injection of a lethal dose of potassium chloride, in accordance with approved protocols. The local governmental review board (Saxon State Directorate) reviewed and approved this study (approval number: TVV 43/2023).

\subsection*{Postprocessing}

Following data collection, we first checked the individual recordings for camera drift and that all data streams had been recorded correctly. Each sequence was then manually reviewed for stability, image quality, and frame rate, and categorised as either usable or discarded. The retained \textit{rosbag} files were decomposed then into files and folders consisting of individual data types (e.g. point clouds, images). 

\begin{figure}[ht]
\centering
\rotatebox{270}{%
\includegraphics[height=0.7\linewidth]{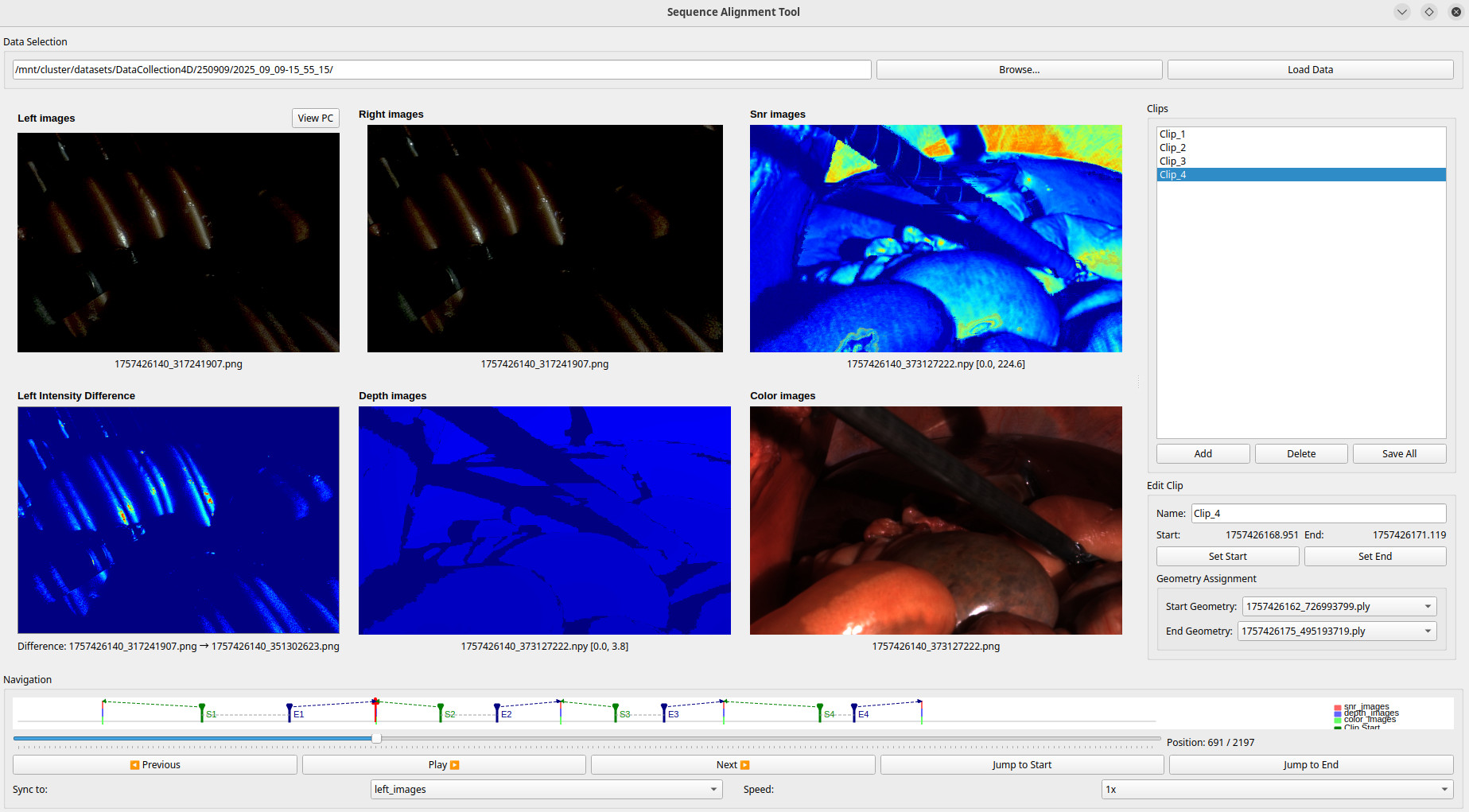}
}
\caption{Sequence Alignment Tool used for creating \texttt{clips.json} files: manually identifying start and end points of clips, and associating them with specific point cloud data.}
\label{fig:Sequencer}
\end{figure}

\begin{figure}[ht]
\centering
\includegraphics[height=0.39\linewidth]{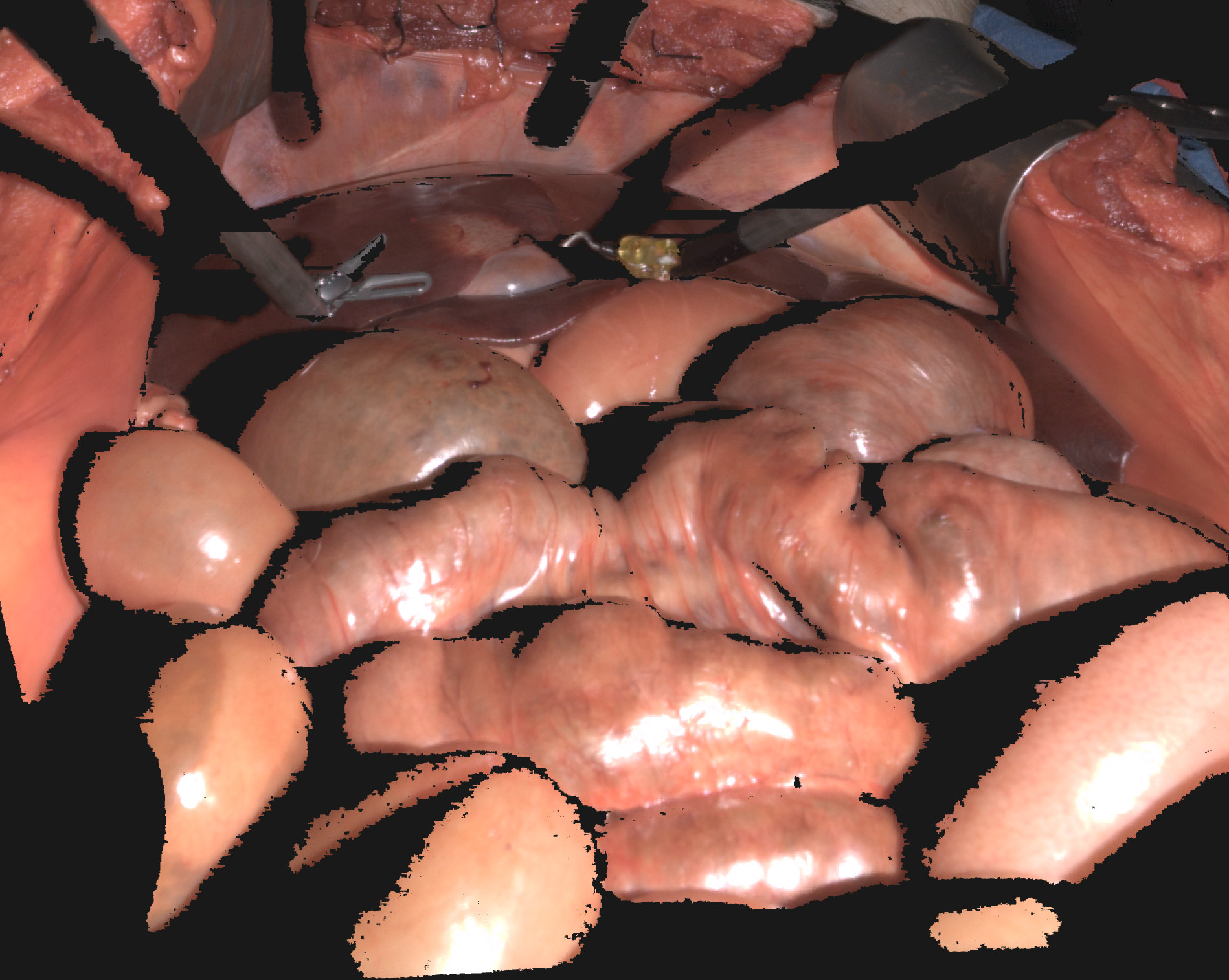}\hfill
\includegraphics[height=0.39\linewidth]{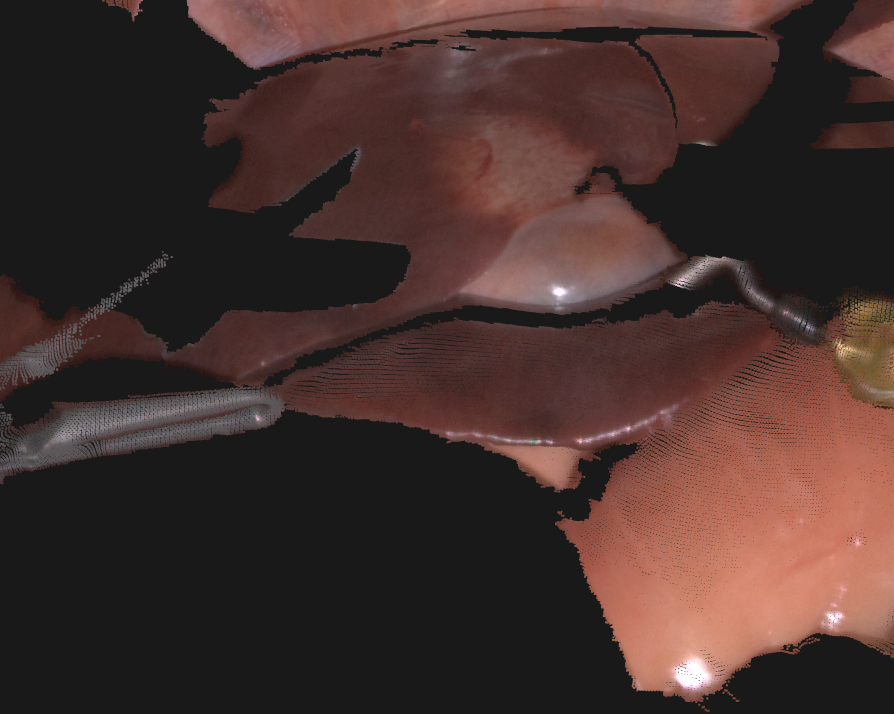}
\caption{Point cloud acquired with the SLC rendered from the perspective of the Zivid camera (left) and the endoscope camera after PnP-RANSAC (right).}
\label{fig:Refinement-pointclouds}
\end{figure}

\begin{figure}[ht]
\centering
\includegraphics[width=\linewidth]{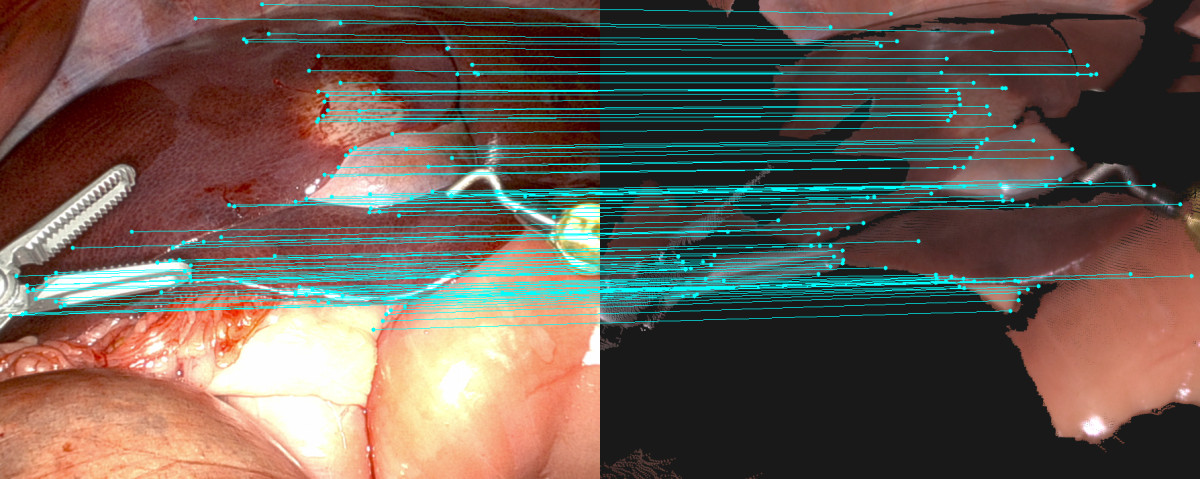}
\caption{Visualization of alignment with feature matching (LightGlue).}
\label{fig:Refinement-matching}
\end{figure}

\begin{figure}[ht]
\centering
\includegraphics[width=0.49\linewidth]{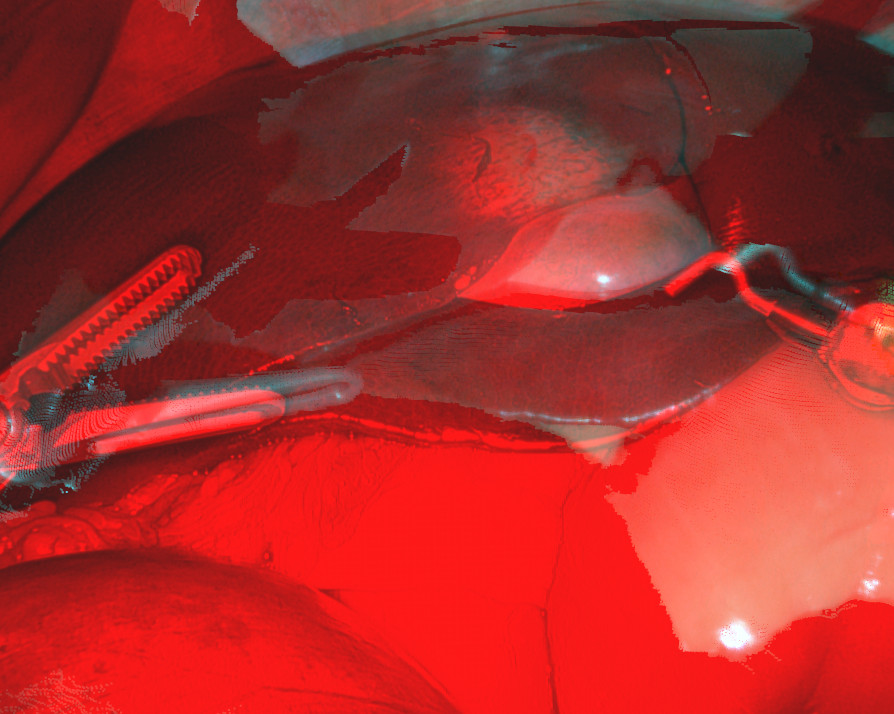}\hfill
\includegraphics[width=0.49\linewidth]{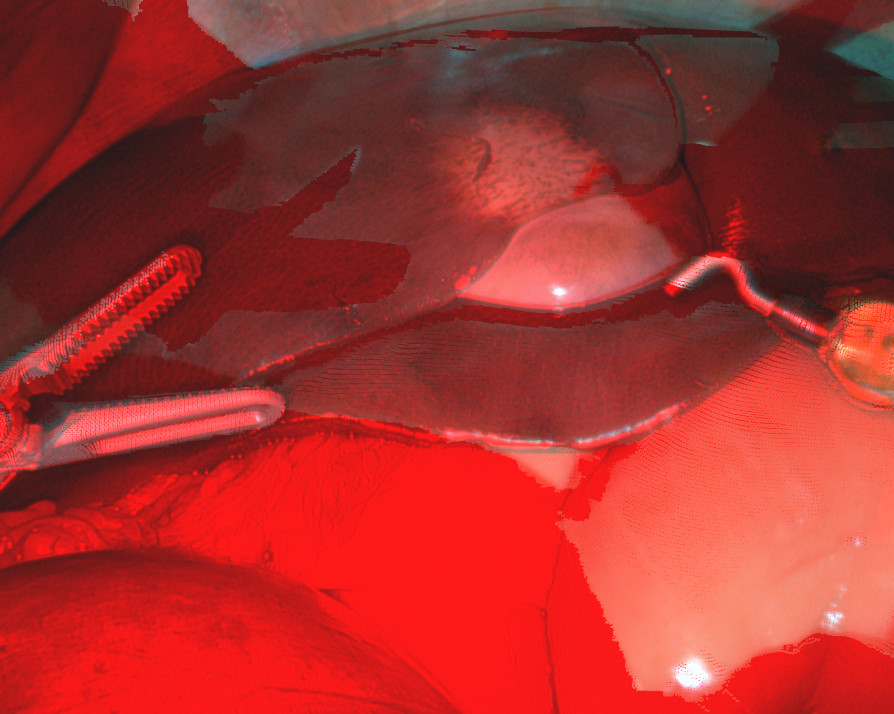}
\caption{Comparison between rendered, aligned, and anaglyph images for nominal (left) and PnP-RANSAC adjusted (right) poses.}
\label{fig:Refinement-anagly}
\end{figure}

\begin{figure}[h!]
\centering
\rotatebox{270}{%
\includegraphics[height=0.7\linewidth]{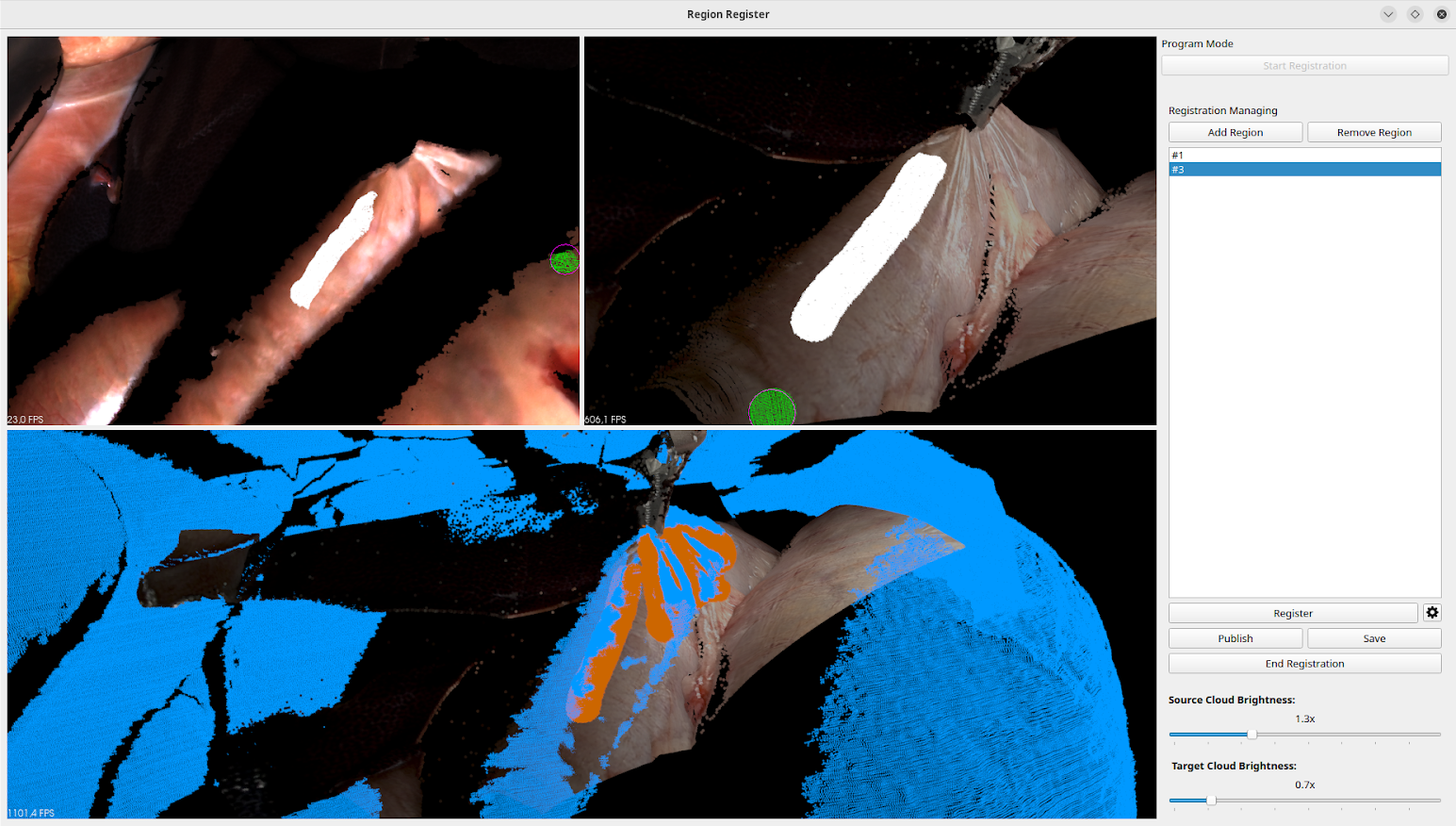}
}
\caption{
Example of the semi-automatic region-based ICP refinement used when standard ICP failed to produce a reliable alignment.
}
\label{fig:Refinement-regionICP}
\end{figure}

During the different recorded segments, before each tissue manipulation, there was a period for which the instruments and tissue were static due to preparing for the manipulation. Additionally, at the end of each tissue manipulation, the scene was not completely static once the instruments stopped moving because the tissues' motion continued for a short time longer (through momentum or sliding). In order for the segments to comprise only the time when the tissues were deforming, a custom graphical tool (Figure~\ref{fig:Sequencer}) was used to annotate the precise start and end points of video segments between successive SLC captures. This was based on visually identifying from the endoscope images the timepoints at which the tissue started and ended moving. Although hand–eye calibration provided approximate pose alignment, residual errors were evident: rendered point clouds from the tracked endoscope perspective did not perfectly coincide with the endoscopic images. To mitigate this, point clouds were reprojected into the camera frame (Figure ~\ref{fig:Refinement-pointclouds} and aligned using feature detection and matching. A corrective transformation was then estimated with PnP+RANSAC, using LightGlue~\cite{lindenberger2023lightglue} as the primary matching algorithm (Figures ~\ref{fig:Refinement-matching} and ~\ref{fig:Refinement-anagly}). In cases where automatic matching failed, correspondences were manually specified. To further refine the poses, Iterative Closest Points (ICP) was carried out between the start and end SLC pointclouds and their stereo depth -derived counterparts. Some ICP registrations did not, however, obtain plausible alignments. As the endoscope camera was almost exclusively static during acquisition in different sessions, each alignment was visually inspected, and the one which appeared most fitting was used for the whole sequence. For sequences where no good alignment was found, a semi-automatic region-based ICP method\cite{clements2008} with manual input was used to obtain higher quality registrations (see Figure \ref{fig:Refinement-regionICP}). We refer to poses which were ultimately retained as 'curated' poses.

Instrument segmentations were also created. For the endoscopic images, segmentations were obtained semi-automatically with the \textit{Segment and Track Anything}\cite{cheng2023segment} and \textit{SAM2}\cite{ravi2025sam2} frameworks. For the SLC data, segmentations were created manually. IGEV++\cite{xu2023igev, xu2025igev++} was used to create depth maps for all of the stereo endoscope image pairs. While all the original data are available at original resolution, these data are all scaled down to a resolution of (\(640\times512\)) for straightforward compatibility with the methods which are evaluated downstream.

Prior to filtering out sequences which were not of high quality, there were 150 sequences collected altogether. After filtering, 98 sequences remained. A summary of the totals can be seen in Table \ref{tab:clip_counts}. Save for the clearly unusable sequences, we are also releasing 30 ambiguous quality sequences in case they still provide some value, but without the substantial post-processing that we performed on the high quality sequences. By default, we intend that the \textit{core} 98 sequences are used for evaluation.

For \emph{Specimen~5}, the left and right images streams were mistakenly swapped during camera calibration, hand-eye calibration and all subsequent recordings. This was corrected in post-processing and all data can be used as for other specimens with the exception of the transforms in the \texttt{tf/} folder (see below), which represent the recorded poses of the \textbf{right} endoscope camera.

\section*{Data Records}

The dataset is organised hierarchically by recording session, capture instance, and manipulation clip.
At the top level, there are specimen folders relating to each pig, each named in the order which data were collected
(i.e., \path{specimen_0}, \path{specimen_1}, …).
Each session folder contains multiple subfolders corresponding to individual capture instances,
named by their recording date and time (e.g., \path{2025_03_06-16_49_40/}).

Within each specimen folder, the following subdirectories and files are provided:

\begin{itemize}
  \item \texttt{camera\_info/}: camera calibration YAML files for endoscope and SLC;
  \item \texttt{clips/}: collection of manipulation episodes, each stored as a separate \texttt{Clip\_X} subfolder (see below);
  \item \texttt{clips.json}: metadata describing clip segmentation and timing;
  \item \texttt{color\_images/}: RGB frames from the structured-light camera (PNG, timestamped);
  \item \texttt{depth\_images/}: depth maps captured by the structured-light camera, timestamped;
  \item \texttt{left\_images/}, \texttt{right\_images/}: raw stereo image pairs prior to rectification;
  \item \texttt{masks/}: mask outputs obtained from SAM-Track\cite{cheng2023segment} and SAM2\cite{ravi2025sam2};
  \item \texttt{pointcloud/}: 3D point clouds obtained from structured-light camera;
  \item \texttt{snr\_images/}: signal-to-noise ratio maps associated with point cloud obtained from structured-light camera;
  \item \texttt{tf/}: transforms, timestamped.
\end{itemize}

Within each \texttt{clips/} directory, individual manipulation episodes are stored as \texttt{Clip\_X} folders. Each clip aggregates the visual streams and geometric captures associated with one manipulation, along with derived products used for downstream reconstruction benchmarking:

\begin{itemize}
  \item \texttt{pose\_bounds.npy}: per-clip pose bounds (identity for static camera, nominal tracked poses for moved camera);
  \item \texttt{curated\_camera\_pose\_\{start,end\}.txt}: nominal camera poses for the start and end SLC captures;
  \item \texttt{left\_images\_rect\_masks/}: per-frame binary masks (filenames are timestamps);
  \item \texttt{left\_images\_rect/}, \texttt{right\_images\_rect/}: rectified left/right images (PNG, timestamped);
  \item \texttt{stereo\_depth/}: per-frame stereo depth (NumPy \texttt{.npy}, timestamped);
  \item \texttt{zivid\_masks/}: Zivid color masks (\texttt{start\_mask.png}, \texttt{end\_mask.png});
  \item \texttt{camera\_info/}: resized camera calibration YAMLs for \(640\times512\) resolution.
\end{itemize}

All rectified images, masks, and depth maps are provided at \(640\times512\) resolution to facilitate consistent benchmarking across reconstruction methods. This hierarchical structure (\texttt{250*/<date\_time>/\{clips/, camera\_info/, ...\}}) mirrors the chronological collection of sessions and ensures reproducible linkage between the raw sensory data, per-clip reconstructions, and calibration metadata.

Lastly, further details, such as the tissue type which is manipulated in each session, is mentioned in \texttt{info.csv}. Whether a sequence is one which features camera movement is indicated in \texttt{session\_details.txt}.

\section*{Technical Validation}
\begin{figure}[h!]
\centering
\includegraphics[width=1.0\linewidth]{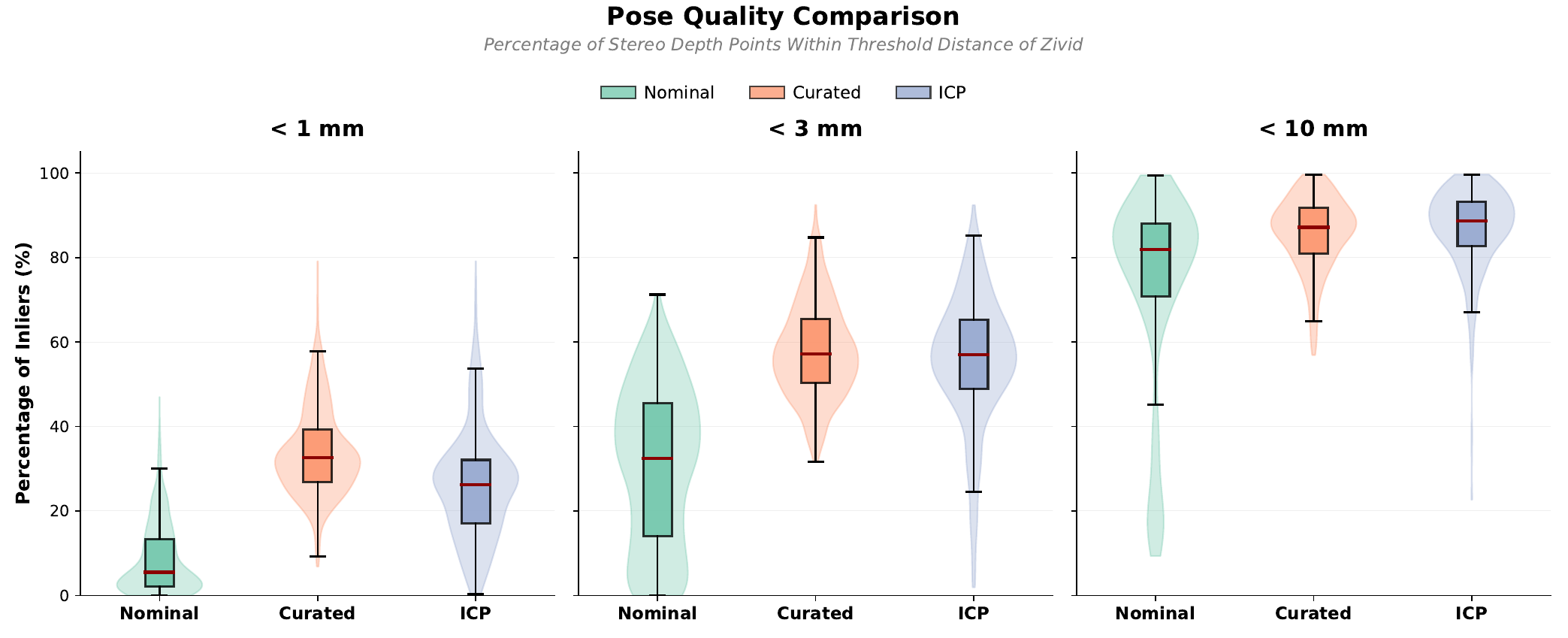}
\caption{
Comparison of pose quality across the Nominal, Curated, and ICP sets. 
The box plots summarize the distribution of nearest–neighbour distances 
between points in the Stereo Depth point cloud and their closest 
corresponding points in the SLC point cloud, illustrating the relative 
accuracy of the poses used in each set.
}
\label{fig:pose_quality_comparison}
\end{figure}

\begin{figure}[h!]
\centering
\includegraphics[width=1.0\linewidth]{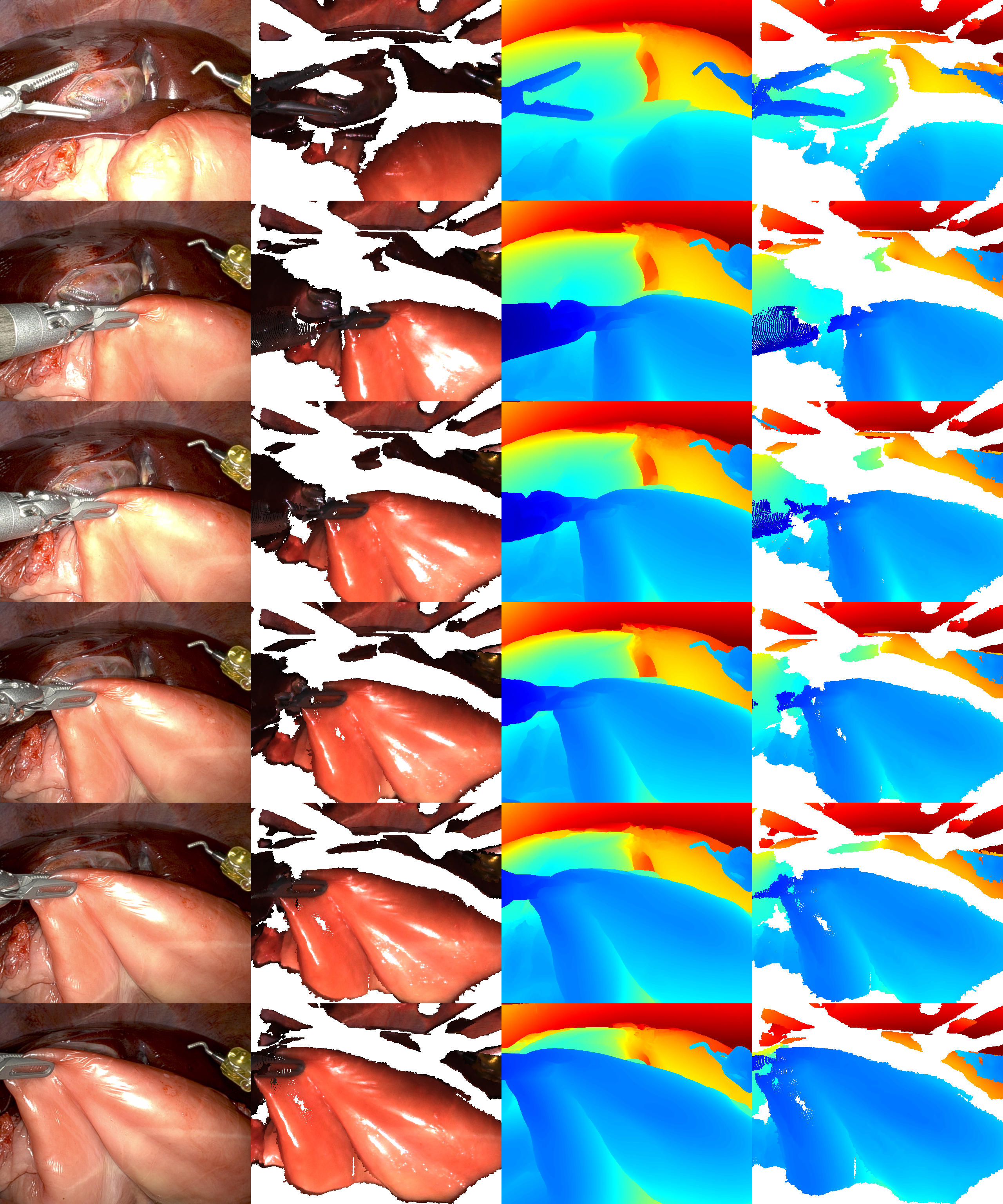}
\caption{
Overview of visual data associated with Clip~1 and the subsequent four clips for sequence \emph{2025\_09\_09-15\_40\_48} of \emph{Specimen~5}. For each clip (from top to bottom: the initial state of Clip~1 followed by the end states of the next four clips), the figure displays: the left endoscopic image, the RGB rendering produced from the SLC point cloud using the curated pose, the stereo depth map, and the SLC point-cloud depth image generated from the same curated pose. This layout illustrates the visual consistency and depth alignment achieved through the curated registration process.
}
\label{fig:specimen5_summary}
\end{figure}

\begin{figure}[htbp]
    \centering
    \begin{minipage}[b]{0.48\textwidth}
        \centering
        \includegraphics[width=\textwidth]{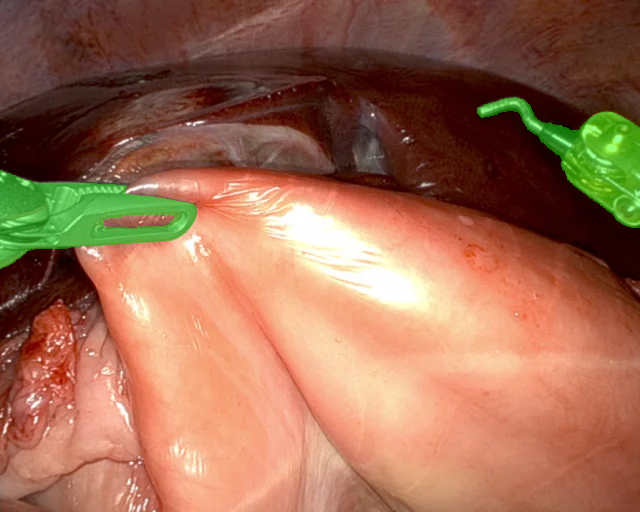}
        \caption*{(a)}
    \end{minipage}
    \hfill
    \begin{minipage}[b]{0.46\textwidth}
        \centering
        \includegraphics[width=\textwidth]{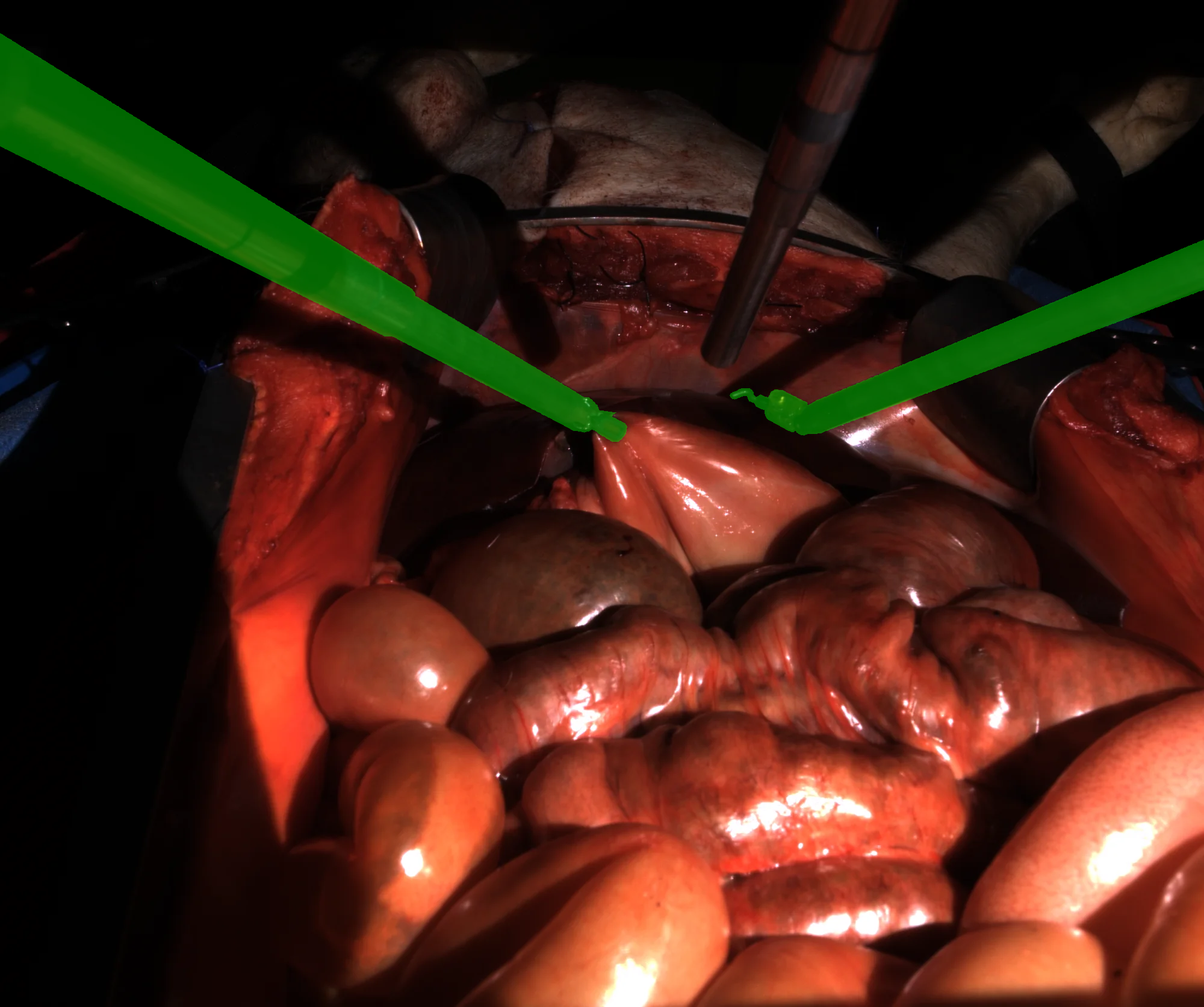}
        \caption*{(b)}
    \end{minipage}
    \caption{Recording \emph{2025\_09\_09-15\_40\_48} of \emph{Specimen~5}, Clip 5 start (i.e., Clip 4 end) images with respective mask overlays: (a) Left endoscope view, (b) Zivid view}
    \label{fig:specimen5-masks}
\end{figure}

To illustrate the quality of the poses obtained through the curation process (including hand-selected ICP-derived and region-based registration poses), we report statistics describing the distance from each point in the Stereo Depth point cloud to its nearest neighbour in the SLC point cloud. These can be found in Table \ref{tab:registration-stats}, with accompanying box-plots in Figure \ref{fig:pose_quality_comparison}. While the ICP registration has more inliers altogether when the threshold is set to 10mm, this changes for 3mm and 1mm thresholds where the curated poses deliver a greater proportion of inliers. This reflects the fact that corresponding points and surfaces between the two pointclouds most tightly align with the curated set of poses. This is further illustrated when comparing the RGB and depth images for the left endoscope images to their equivalents rendered from curated poses using the SLC pointclouds, as can be seen in Figure \ref{fig:specimen5_summary}. Examples of the accompanying segmentation masks can be seen in Figure \ref{fig:specimen5-masks}.

\begin{figure}[ht]
\centering
\includegraphics[width=0.49\linewidth]{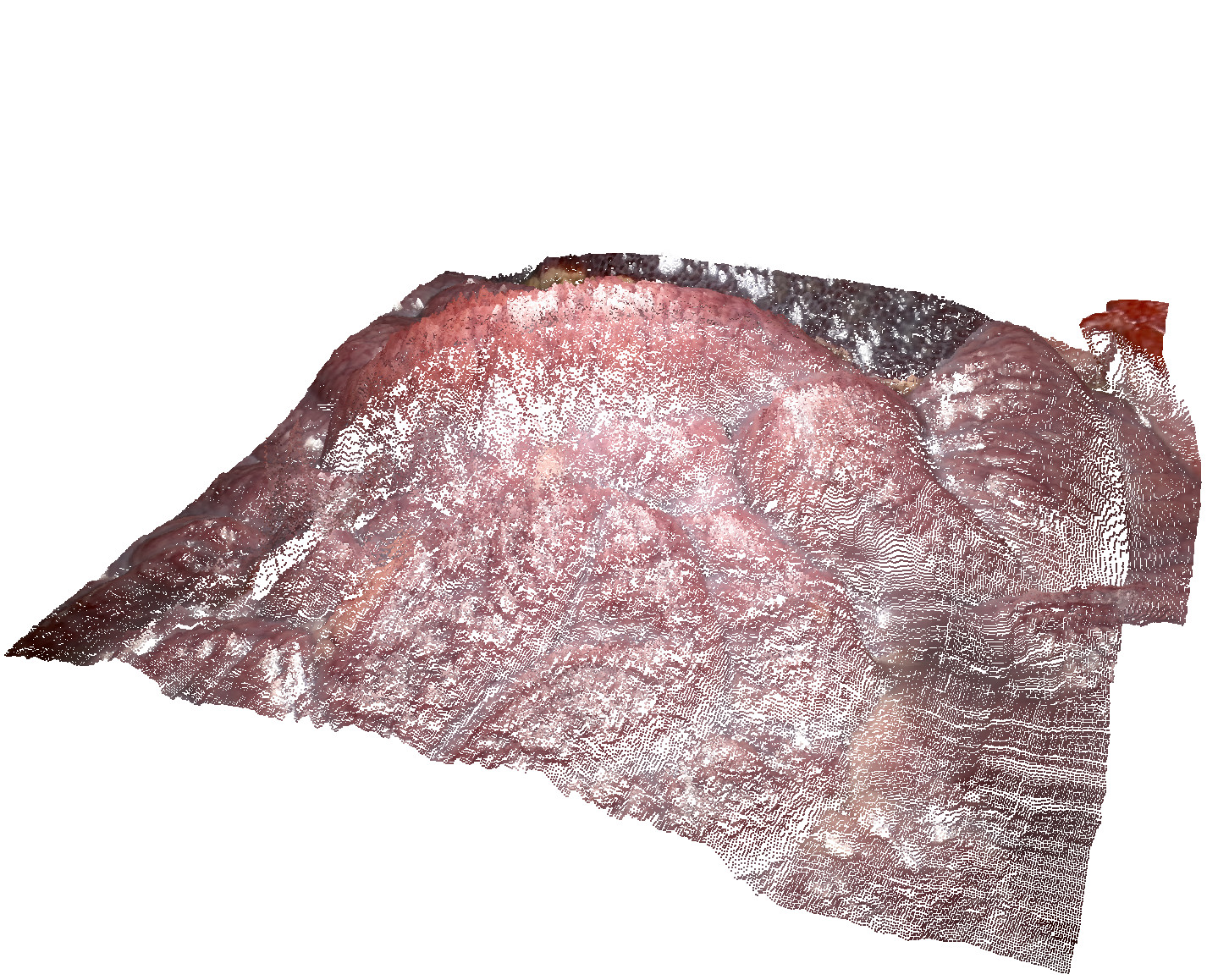}\hfill
\includegraphics[width=0.49\linewidth]{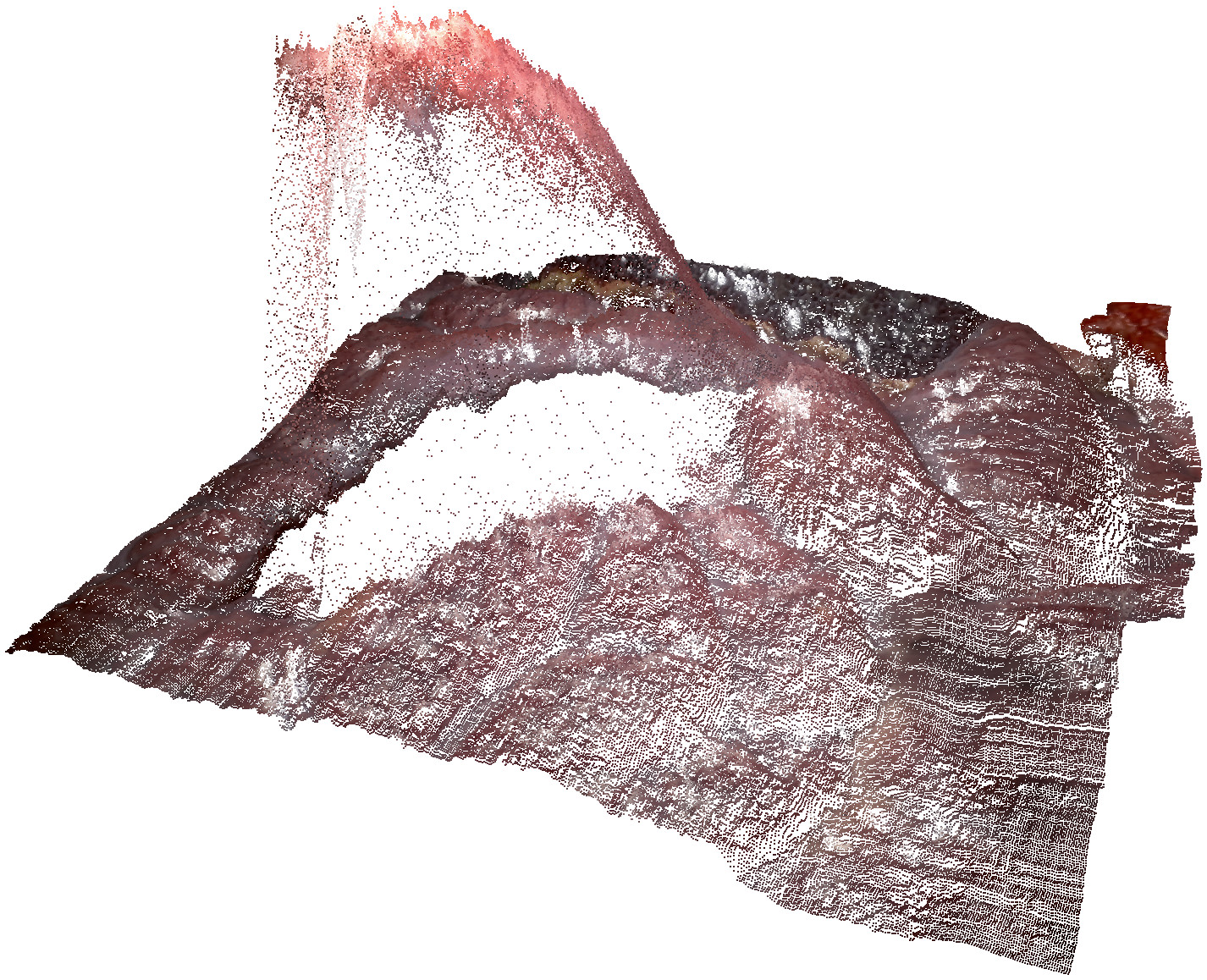}
\caption{Sample pointclouds extracted for initial (left) and end (right) states on one sequence from our dataset, obtained using the ForPlane\cite{yang2024forplane} method.}
\label{fig:sample-reconstruction}
\end{figure}

As we intend for the dataset to be used to evaluate 4D Non-Rigid reconstruction in minimally-invasive abdominal surgery, we furthermore show the results of 3D reconstruction for the ForPlane Method\cite{yang2024forplane} on one of our sequences. A point cloud extracted for the initial and final states is obtained and displayed, as visible in Figure \ref{fig:sample-reconstruction}.

\section*{Usage Notes}
This dataset supports research on non-rigid 3D/4D reconstruction, SLAM and depth estimation in minimally invasive surgery. Use rectified images, masks, and stereo depth in \texttt{clips/}, intrinsics in \texttt{camera\_info/}, and \texttt{curated\_camera\_pose\_\{\-start,end\}.txt} to align reconstructions to structured-light start/end point clouds. Evaluate geometry by comparing to SLC point clouds and optionally assess photometric quality of view-synthesis with PSNR/SSIM/LPIPS on non-masked pixels. Minor residual pose misalignments may remain despite the pose refinement steps undertaken. 

\textit{Whole} sequences assess end-to-end robustness, \textit{Incremental} sequences probe deformation magnitude and enable a more granular study of tissue deformation, and \textit{Moved-camera} sequences test handling of out-of-view deformation and camera movement. While the \textit{Moved-camera} sequences test the reconstruction of out-of-view deformation in a more significant manner, all sequences in fact possess out-of-view deformation whose evaluation this dataset also aims to enable. The ideal ground truth for this assessment would be to know correspondences between the points in the SLC pointclouds and points in the reconstructed pointclouds derived from the clips themselves - this is in practice, however, difficult to achieve. A lower-hanging fruit would be to evaluate simply what lies behind masked out tools, assuming that the respective tools have moved and what lies behind has already been observed during the clip. 

If the evaluation protocol involves tuning hyperparameters (or training parameters) on the dataset prior to testing, we recommend using a 4:1 train–test split (across the 98 \textit{core} sequences of Specimens 1 to 5) within a 5-fold Leave-One-Out cross-validation setup.

The data are usable with standard open-source Python tools and can be quickly interacted with through the repository below. Please cite this Data Descriptor when using the dataset.

\section*{Data Availability}
The D4D dataset is publicly available via OPARA at \url{https://doi.org/10.25532/OPARA-1033}. The dataset includes rectified stereo images, instrument masks, stereo depth maps, structured-light point clouds, curated camera poses, and camera intrinsics for 98 curated recordings across six data collection sessions.

\section*{Code Availability}
For ease of use, we are pairing the dataset with a repository which makes loading the data convenient and simple. The repository can be found at: \url{https://github.com/reubendocea/d4d}.

\section*{Author Contributions}

R.D. conceived the study, developed the technical setup, led data collection, carried out data handling \& post-processing. M.F. principally developed the semi-automatic region register tool for aligning pointclouds. J.X., C.L., and J.B. assisted during data collection. As surgical staff, R.Y., A.S., and M.W. contributed to data collection by helping in setup and operation of the da Vinci surgical robot. Y.L. gave regular feedback on data quality and usefulness, making valuable suggestions. M.W. defined surgically realistic movement sequences. Y.L., Q.D., M.P., M.W. and S.S. assisted in study conception, provided methodological input and supervision. All authors reviewed the manuscript.

\section*{Competing Interests}

The authors declare no competing interests.

\section*{Acknowledgements}

We thank the team of the Experimental Operating Room at the NCT Dresden for their support.

\section*{Funding}

This project has received funding from the European Union's Horizon Europe research and innovation programme under grant agreement No 101092646. This work was supported by the German Research Foundation (DFG, Deutsche Forschungsgemeinschaft) as part of Germany's Excellence Strategy – EXC 2050/1 – Project ID 390696704 – Cluster of Excellence "Centre for Tactile Internet with Human-in-the-Loop" (CeTI) as well as by the German Federal Ministry of Health (BMG) within the SurgOmics project (grant number BMG 2520DAT82). The authors acknowledge the financial support by the Federal Ministry of Research, Technology and Space of Germany in the programme of "Souverän. Digital. Vernetzt.". Joint project 6G-life, project identification number: 16KISK001K. This work is supported by the project "Next Generation AI Computing (gAIn)," funded by the Bavarian Ministry of Science and the Arts and the Saxon Ministry for Science, Culture, and Tourism. The authors acknowledge the financial support by the Federal Ministry of Research, Technology and Space of Germany in the programme of “DigiLeistDAT”. Joint project SurgicalAIHubGermany, project identification number: 02K23A112. 

\section*{Figures \& Tables}

\begin{table}[h!]
\centering
\caption{Summary of Clip Counts by Specimen}
\label{tab:clip_counts}
\begin{tabular}{
    l
    S[table-format=3.0]
    S[table-format=2.0]
    S[table-format=2.0]
    S[table-format=2.0]
    S[table-format=2.0]
}
\toprule
& \multicolumn{1}{c}{\textbf{Pre-Filter}} & \multicolumn{4}{c}{\textbf{Retained}} \\
\cmidrule(lr){2-2} \cmidrule(lr){3-6}
\textbf{Specimen} & \textbf{Prior to Filter} & \textbf{Total} & \textbf{(i) Single} & \textbf{(ii) Incremental} & \textbf{(iii) Moved Camera} \\
\midrule
0 & 24 & 0 & 0 & 0 & 0 \\
1 & 37 & 16 & 9 & 7 & 0 \\
2 & 14 & 14 & 7 & 5 & 2 \\
3 & 19 & 15 & 8 & 6 & 1 \\
4 & 33 & 32 & 15 & 12 & 5 \\
5 & 23 & 22 & 10 & 9 & 2 \\
\midrule
\textbf{TOTAL} & \textbf{150} & \textbf{98} & \textbf{49} & \textbf{39} & \textbf{10} \\
\bottomrule
\end{tabular}
\end{table}

\begin{table}[h!]
\centering
\small
\begin{tabular}{lcccccc}
\toprule
\textbf{Metric / Set} & \textbf{Mean} & \textbf{Median} & \textbf{Std} & \textbf{Min} & \textbf{Max} \\
\midrule

\textbf{Median Dist (mm) - Nominal} & 5.63 & 4.25 & 3.51 & 1.10 & 15.25 \\
\textbf{Median Dist (mm) - Curated} & 2.37 & 2.15 & 1.20 & 0.52 & 7.26 \\
\textbf{Median Dist (mm) - ICP}     & 2.70 & 2.32 & 1.69 & 0.52 & 14.11 \\
\midrule

\textbf{Inliers $<$1mm (\%) - Nominal} & 8.60 & 5.48 & 8.40 & 0.00 & 46.98 \\
\textbf{Inliers $<$1mm (\%) - Curated} & 33.90 & 32.62 & 10.38 & 6.84 & 79.11 \\
\textbf{Inliers $<$1mm (\%) - ICP}     & 26.02 & 26.16 & 12.85 & 0.26 & 79.13 \\
\midrule

\textbf{Inliers $<$3mm (\%) - Nominal} & 30.65 & 32.46 & 18.46 & 0.00 & 71.22 \\
\textbf{Inliers $<$3mm (\%) - Curated} & 57.89 & 57.22 & 11.19 & 31.58 & 92.45 \\
\textbf{Inliers $<$3mm (\%) - ICP}     & 55.74 & 56.95 & 15.34 & 1.92 & 92.42 \\
\midrule

\textbf{Inliers $<$10mm (\%) - Nominal} & 73.56 & 81.86 & 23.74 & 9.33 & 99.47 \\
\textbf{Inliers $<$10mm (\%) - Curated} & 85.71 & 87.16 & 8.63  & 56.91 & 99.62 \\
\textbf{Inliers $<$10mm (\%) - ICP}     & 86.68 & 88.63 & 10.15 & 22.64 & 99.64 \\
\bottomrule
\end{tabular}
\caption{Summary statistics of point-to-point distances from points in Stereo Depth -derived point cloud to SLC pointcloud after transformation with Nominal, Curated, and ICP poses evaluation sets (N = 540 pointclouds).}
\label{tab:registration-stats}
\end{table}

\end{document}